\documentclass[letterpaper, 10 pt, conference]{ieeeconf}  % Comment this line out if you need a4paper

\usepackage{times}
\usepackage{epsfig}
\usepackage{amsmath}
\usepackage{amssymb}
\usepackage{graphicx}
\usepackage{bm}
\usepackage{caption}
\usepackage{float}  %设置图片浮动位置的宏包
\usepackage{subfigure}
\usepackage{graphbox}
\usepackage{ amssymb }
\usepackage{multirow}
\usepackage{hyperref}
\makeatletter
\def\hlinew#1{%
  \noalign{\ifnum0=`}\fi\hrule \@height #1 \futurelet
   \reserved@a\@xhline}
\makeatother
\IEEEoverridecommandlockouts                              % This command is only needed if 
\title{\LARGE \bf
MVM3Det: A Novel Method for Multi-view Monocular 3D Detection
}

\author{Haoran Li$^{1}$, Zicheng Duan$^{2}$, Mingjun Ma$^{1}$, Yaran Chen$^{1}$, Jiaqi Li$^{3}$ and Dongbin Zhao$^{1}$% <-this % stops a space
\thanks{$^{1}$Haoran Li, Mingjun Ma, Yaran Chen and Dongbin Zhao are with the State Key Laboratory of Management and Control for Complex Systems, Institute of Automation, Chinese Academy of Sciences, Beijing, 100190, China, and also with the University of Chinese Academy of
Sciences, Beijing, China (email : lihaoran2015@ia.ac.cn,  mingjun.ma@ia.ac.cn, chenyaran2013@ia.ac.cn, dongbin.zhao@ia.ac.cn)}%
\thanks{$^{2}$Zicheng Duan is with the College of Engineering and Computer Science, Australian National University, ACT, 2601, Australia (email : zicheng.duan@anu.edu.au)}%
\thanks{$^{3}$Jiaqi Li is with the School of Mechanical Engineering, Beijing Institute of Technology, Beijing, 100081, China (email: xuer0324@gmail.com)}
\thanks{This work is supported by the Strategic Priority Research Program of Chinese Academy of Sciences (CAS) under Grant XDA27030400, 
and the National Natural Science Foundation of China (NSFC) under Grants No. 62006226. }
}
\begin{document}

\maketitle
\thispagestyle{empty}
\pagestyle{empty}

%%%%%%%%%%%%%%%%%%%%%%%%%%%%%%%%%%%%%%%%%%%%%%%%%%%%%%%%%%%%%%%%%%%%%%%%%%%%%%%%
\begin{abstract}
% 3D object detection is vital for autonomous driving and robotic industries. 
% Despite extensive researches on monocular 3D object detection, 
% occlusion has always been a nonnegligible gap between laboratory research and industrial applications. 
% To address the occlusion problem in 3D detection, we propose a novel multi-view 3D detection algorithm. 
% We apply feature map projection to aggregate multi-cue information to obtain localization information in birds-eye-view and estimate orientation in normal perspectives. 
% We test our model on WildTrack pedestrian detection public dataset for localization performance evaluation and achieved ???\% MODA, 
% which is slightly inferior to the current state-of-the-art method. 
% We also verify our method in real-life scenarios by applying the algorithm on the ICRA RoboMaster dataset, 
% and the model achieves 96.2\% MODA in localization and 83.5\% AOS with respect to orientation estimation.

Monocular 3D object detection encounters occlusion problems in many application scenarios, such as traffic monitoring, pedestrian monitoring, etc., 
which leads to serious false negatives. Multi-view object detection effectively solves this problem by combining data from different perspectives. 
However, due to label confusion and feature confusion, the orientation estimation of multi-view 3D object detection is intractable, 
which is important for object tracking and intention prediction. 
In this paper, we propose a novel multi-view 3D object detection method named MVM3Det which simultaneously estimates the 3D position and orientation of the object 
according to the multi-view monocular information. 
The method consists of two parts: 1) Position proposal network, 
which integrates the features from different perspectives into consistent global features through feature orthogonal transformation to estimate the position. 
2) Multi-branch orientation estimation network, 
which introduces feature perspective pooling to overcome the two confusion problems during the orientation estimation. 
In addition, we present the first dataset for multi-view 3D object detection named MVM3D\footnote{We will release the code and dataset at \url{https://github.com/DRL-CASIA/MVM3D}.}. 
Comparing with State-Of-The-Art (SOTA) methods on our dataset and public dataset WildTrack, our method achieves very competitive results.
\end{abstract}

%%%%%%%%%%%%%%%%%%%%%%%%%%%%%%%%%%%%%%%%%%%%%%%%%%%%%%%%%%%%%%%%%%%%%%%%%%%%%%%%
\section{INTRODUCTION}

Object detection is a fundamental task for autonomous robot system, which has developed vigorously and has been widely deployed in perception systems. 
However, due to the lack of 3D information, 2D object detection is hard to provide sufficient information for decision-making and planning of autonomous robots. 
Therefore, a large amount of 3D object detection methods have been developed for 3D pose estimation, 
such as monocular-based methods \cite{oftnet,yolo6d}, 
LiDAR-based methods \cite{voxelnet,prcnn,8908931}, and multi-sensor fusion methods \cite{mv3d,avod,chen2020boost}. 
However, these methods only use the sensor data obtained from a single perspective, and the mutual occlusion between objects causes serious information loss. 
Even for multi-sensor fusion methods whose relative positions between the sensors are close, 
the ability of these methods to alleviate blind spots in the field of view is limited. Since almost current methods are based on a single view, 
the false negative caused by the occlusion has become one of the main problems in the application.

% Utilizing multiple sensors from different perspectives would be one of the possible solutions for detections under occlusion. 
% Unlike multi-sensor fusion detection, which refers to multiple sensors from a single perspective, 
Multi-view 3D object detection is one of the possible solutions for this problem,
which fuses the information from different perspectives to overcome detection failures 
caused by occlusion in single perspective scenarios, hence providing robust results and accurate pose estimation.
However,
estimating a unified 3D pose from sensor data captured from different perspectives is the main challenge of multi-view object detection, 
especially in the case of lack of depth information. 
Firstly, since the information from each view is different,  
how to effectively fuse multi-view information into unified global information is inevitable problem for 3D location estimation. 
% \cite{deepocclusion} fuses 2D detection results in each view 
% and uses the conditional random field to estimate objects global position. 
% Hou \emph{et al.} proposes MVDet \cite{mvdet} which introduces feature map projection to fuse information and designs large convolution kernels to aggregate spatial information. 
Secondly, % orientation estimation with multiple viewing information is also very difficult under multiple monocular cameras.
orientation is vital for trajectory tracking and intention prediction, 
but orientation estimation with multiple viewing information is intractable with two confusion problems:
% There are two confusion problems which makes orientation estimation intractable: 
1) label confusion and 2) feature confusion.
The first problem is common in monocular 3D detection. Relative orientation labels are usually used to eliminate label ambiguity caused by different line of sight angles. 
However, since the relative orientation labels in each view are different, 
it is difficult to find a consistent relative orientation label for all different views in multi-view detection. 
The second problem is that due to the perspective transformation of multiple cameras, the global features of objects which have same orientation and different positions are different. 
At present, there is little work focusing on overcoming this feature confusion for accurate orientation estimation.

% Datasets are essential for learning based methods. 
Another important factor restricting the development of multi-view 3D object detection methods is the lack of datasets.
At present, although there are many datasets \cite{Geiger2013IJRR,waymo_open_dataset} for monocular 3D object detection, 
these datasets are only limited to verifying the detection methods from a single perspective. 
Arnold \emph{ et al.} \cite{9228884} uses LiDAR to collect roundabout data in the simulation environment to develop LiDAR-based multi-view 3D object detection method, 
but does not involve multi-view monocular data. The WildTrack dataset \cite{wildtrack} is closest to the requirements of multi-view monocular 3D detection method. 
However, this dataset not only contains a small amount of data, but also has no orientation labels of objects.
Therefore, the development of multi-view monocular 3D object detection method is largely limited to the lack of appropriate datasets.

In this paper, we propose a novel multi-view monocular 3D object detection method MVM3Det that simultaneously estimates the position and 
orientation. %% using the multi-view sensor data hence alleviate the false detection problem caused by occlusion.
This method contains position proposal network which achieves the consistent global features through feature orthogonal projection, 
and  multi-branch orientation estimation network which alleviates features confusion problem with feature perspective pooling. 
In addition, we present the dataset MVM3D for multi-view 3D object detection, hoping to promote the development of multi-view 3D object detection methods. 
The main contributions of this paper are as follows.
\begin{itemize}
   % \item[1)] A multi-view 3D object detection method is proposed, which effectively alleviates the false negative problem caused by the occlusion and
   %  estimates precise localization and object orientation in occluded scenarios.
   \item[1)] In order to estimate unique position of the object with multi-view information,
    we propose feature orthogonal projection and construct a position proposal network. 
    This network generates the consistent global features from multi-view cues. 
   \item[2)] 
   We design multi-branch orientation estimation network with feature perspective pooling to estimate the corresponding relative orientation for each perspective.
   This network alleviates features confusion problem during orientation estimation. 
   Based on above modules, a multi-view 3D detection method is proposed.
   It is worth mentioning that our method is the first method that estimates the orientation and position simultaneously 
   with multi-view monocular information. 
   \item[3)]
   We present the first multi-view 3D object detection dataset named MVM3D.
   Compared with the SOTA methods on our dataset and public dataset WildTrack, the proposed method achieves very competitive results. 
   On MV3D dataset, our method achieves 95.9\% MODA, an 1.1\% increases over SOTA method.
   Our method achieves 49.0\% $AP_{3D}$, an 9.7\% increases over monocular based 3D detection method.
   % As for localization performance, our method achieves 96.2\% MODA, which surpasses the state-of-the-art result\cite{mvdet} 94.8\% MODA on the RoboMaster Dataset by 1.4\%. 
   % For orientation estimation, we achieves 71.2\% AOS and 75.4\% AP, outperforms our monocular feature projection baseline results 55.7\% AOS and 60.3\% AP.
\end{itemize}

\section{REALTED WORK}
% In this section, we summarize existing 2D detection and mono/multi-sensor 3D detection methods and multi-view detection methods.
% \subsection{2D object detection methods}
% 2D image-based detection algorithms\cite{rcnn, fastrcnn, fasterrcnn, ssd, yolo, yolov2, yolov3} has been greatly developed.
% Yolo family\cite{yolo,yolov2,yolov3} regards object detection as a regression problem, these methods takes RGB images as input and directly output the localization and classification result with the help of pre-defined grids. 
% Regarding two-stage detection, Faster RCNN\cite{fasterrcnn} groundbreakingly presents region proposal network (RPN), which relieves the heavy workload from selective search region proposal methods in \cite{rcnn,fastrcnn}. RPN are since widely used in various algorithms as a concise and effective way of extracting RoI, such as in \cite{mv3d}. We also utilize RPN in our work to propose region of interests.
% This section reviews the related work of 3D object detection in recent years, including monocular based methods and multi-view based methods.

\subsection{Monocular based detection}
% There are two kinds of 3D object detection methods based on single view: image-based method and point cloud method. 
There are serveral monocular based 3D detection methods for autonomous robots.
Due to the lack of depth information in the image, some assumptions are usually required to restore the 3D pose of the object from the image. 
For example, objects are assumed to have the same altitude \cite{mono3d}. This assumption cannot be satisfied in many automatic driving scenes. 
Therefore, OFTNet \cite{oftnet} realizes orthogonal feature transformation within a certain height range to weaken the previous assumption. 
In addition, some methods \cite{geometry, liu2019deep, gs3d} abandon the altitude hypothesis. They use the 2D bounding boxes obtained by the 2D object detection method, 
and predict the size and orientation of the object. 
Finally, the spatial position of the object is estimated by solving perspective N – points problem. These methods are sensitive to the prediction accuracy of object size and orientation. 
Another kind of methods is based on monocular depth estimation \cite{roi10d, monopsr, monogrnet}. This kind of methods needs to pre-train the depth estimation network, 
and then estimate the 3D pose of the object combined with the 2D detection frame.

\subsection{Multi-view based detection}
% Either the single-view based methods or multi-sensor fusion based methods cannot overcome the problem of false negative caused by occlusion. 
% This phenomenon is more serious when objects are crowded.
Monocular based methods confront the false negative problem caused by the occlusion, especially when the objects are crowded. 
In the past two years, some multiple perspectives based methods have been developed to overcome the occlusion problem by capturing information from multiple perspectives.
Fleuret \emph{et al.} \cite{pom2008} propose probabilistic occupancy map method which estimates the probabilities of occupancy on the ground plane with multi-view information. 
Peng \emph{et al.} \cite{ PENG20151760} builds a multi-view network which combines multiple Bayesian networks from per view and predicts the localization of the objects. 
Baque \emph{et al.} \cite{deepocclusion} fuses the results of 2D target detection and used convolutional neural network and conditional random field to estimate pedestrian occupancy jointly. 
Hou \emph{et al.} \cite{mvdet} chooses an efficient anchor free feature aggregation method to regress pedestrian occupancy as a Gaussian distribution.
However, these methods only realize position estimation for pedestrian detection, and do not estimate the orientation of objects. 
In this paper, a novel multi-view 3D object detection method is proposed, which estimates the position and orientation simultaneously.

% Multi-view detection is one of the new ideas for 3D object detection. 
% Multiple sensors from different perspectives can be used to compensate for the limitation of the field-of-view under a single perspective hence providing occlusion-free information. 
% At present, multi-view detection has been widely researched for person re-identification and vehicle re-identification tasks. 
% However, the purpose of ReID is not compatible with normal detection tasks. 
% In recent years, multi-view overlap area detection algorithms have achieved remarkable results in occluded pedestrian detection scenarios. 
% For example, \cite{deepocclusion} fused the results of 2D target detection and used CNN and Conditional Reference Field (CRF) to estimate pedestrian occupancy jointly. 
% While \cite{mvdet} choose an efficient anchor free feature aggregation method to regress pedestrian occupancy as a Gaussian distribution. 
% Its pioneering feature projection and aggregation method provide a new starting point for utilizing multi-view features.

\begin{figure*}[htbp]
   % \hspace{0.5cm}
   \centering
   \scalebox{0.9}{
   \includegraphics[width=1\textwidth]{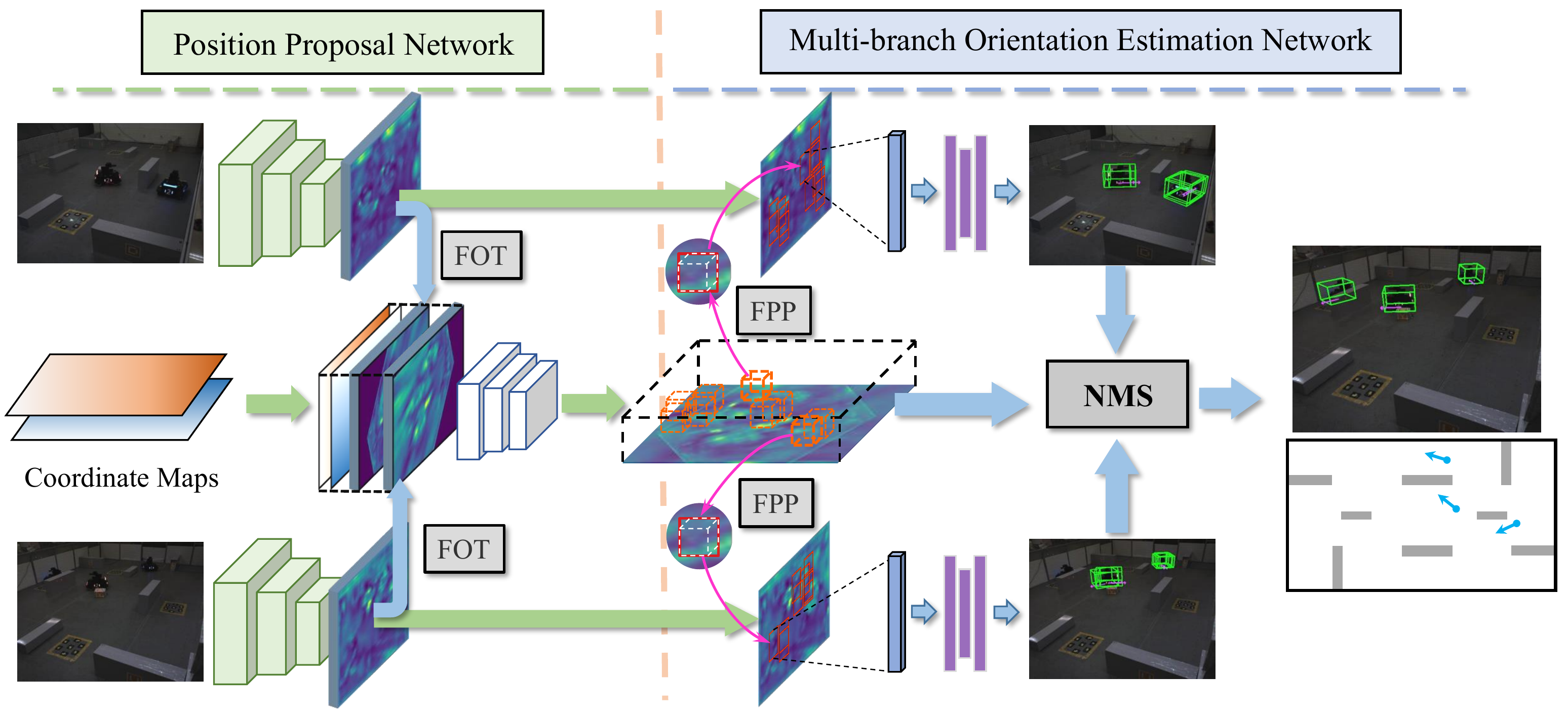}
   }
   \caption{The architecture of MVM3Det. 
   The network consists of PPN and multi-branch orientation estimation network. 
   PPN estimates the position of the object by fusing the information from multiple perspectives through the Feature Orthogonal Transformation (FOT). 
   Multi-branch orientation estimation network predicts the orientation in each perspective through the Feature Perspective Pooling (FPP).
   Non-Maximum Suppression (NMS) is used to select final prediction.} 
   \label{fig_pipeline}
   \vspace{-0.5cm}
\end{figure*}
\section{Method}
% We propose a multi-view network to conduct occlusion-free 3D detection, we define the 3D detection problem as a object pose estimation problem, 
% where pose refers to object location and orientation. The network takes multi-view images as input and outputs object localization and accurate orientation. 
% The network consists of three main modules: multi-view feature fusion network, position proposal network, and pooling-based orientation estimation network. 
% The multi-view fusion network takes images as inputs and outputs the projected birds-eye-view (birds-eye-view) feature map, 
% while the position proposal network provides rough location estimation and corresponding rough 3D boxes based on birds-eye-view features. 
% In the following multi-view RoI Pooling module, 3D boxes are projected into per-view 2D boxes for orientation estimation and position proposal fine-tuning. 
% These modules are introduced in the following sections respectively.

In this paper, we propose a novel multi-view 3D object detection method. This method simultaneously estimates 3D position and orientation from multi-view monocular information, 
and effectively alleviate the problem of false negative caused by the occlusion. 
This method mainly consists of two parts: position proposal network and multi-branch orientation estimation network. 
Position proposal network obtains the consistent global features from data of different perspectives, and uses anchor-based method to estimate the spatial position of objects. 
Multi-branch orientation estimation introduces feature perspective pooling to realize orientation estimation from different perspectives according to the previous position, 
so as to alleviate the confusion problem. The network structure is shown in Fig. \ref{fig_pipeline}.

\subsection{Position proposal network}
% Fusion of sensor information from different perspectives to obtain a consistent global feature is the key to achieving multi-view 3D object detection. In our work, this part is divided into two modules: feature extraction and feature-level fusion. Feature extraction refers to using deep convolutional networks to extract feature information conducive to 3D detection from multiple sensor data sources. This paper takes camera images as examples. A Resnet-18 backbone is used to extract features from different image spaces. 

The purpose of Position Proposal Network (PPN) is to estimate the possible 3D spatial position of objects according to the data obtained from different perspectives. 
Different from the classical Region Proposal Network (RPN), PPN includes three processes: feature extraction, feature fusion and position proposal. 
In this paper, the monocular camera is used for each view to monitor the objects. Considering the great breakthrough of deep convolutional networks in the field of image processing, 
we employ ResNet-18 \cite{resnet} as the feature extractor of the image in each perspective to capture deep features.

% The information from multiple image spaces is different, and the positions of the same object from different perspectives also vary. Therefore, aligning the extracted features maps from multiple image spaces to a unified feature space is necessary. In this paper, we project image feature maps from multiple sources into a birds-eye-view space and stack the birds-eye-view features vertically as the final unified feature map. The projection of the conversion process is as follows:
% \subsubsection{Feature orthogonal transformation}
Since the information observed from each perspective is different, it is meaningless to directly fuse the deep features obtained by ResNet in image space. 
Therefore, the premise of feature fusion is to align the features captured from different perspectives in a unified feature space. 
In this paper, we assume that the altitude of the objects is approximately distributed on a horizontal plane, 
then we introduces the {\bf feature orthogonal transformation} to project the deep features from different perspectives into a Birds-Eye-View (BEV) space.
Similar to \cite{oftnet,mvdet,li2021bifnet}, when the altitude of the distribution plane of the objects is known, 
we calculate the position of each pixel from deep features in the BEV space by orthogonal transformation.
$$
% F_{bev} = f(P \cdot F_{image})
\Gamma = \tau (R^{-1} K^{-1} U + R^{-1} T).
$$
Here $\Gamma$ is 3D position, and $U$ is the homogeneous coordinates of the pixel in the camera image.
$K$ is the camera matrix. $R$ and $T$ are the rotation matrix and translation vector of the camera, respectively.
$\tau$ is scale factor which is estimated by 
$$
\tau = \frac{(R^{-1} K^{-1} U + R^{-1} T)|_z}{z^{P}}.
$$
$V|_z$ means the $z$ axis value of the vector $V$. $z^P$ is the altitude of the objects.
For each deep feature of each view, the above orthogonal transformation is used to convert image features to the BEV space. 
We stack the projected features with coordinate maps and fuse them in the BEV space by the convolutional network, so as to obtain consistent global features.
% where $f$ represents feature map aggregation, $P$ denotes projection matrix, and $F_{bev}$ means birds-eye-view feature map. Besides feature map projection and aggregation, image projection followed by feature extraction is an alternative projection option. However, image projection will break the relationship between pixels and distort spatial information, leading to inaccurate detections. Results for different projection options are demonstrated in TABLE III.

Based on the global features, we introduce anchor-based method for location estimation. 
Similar to RPN, the global feature is input into the fully convolutional network to predict the probability of an object at the corresponding position of each anchor and 
the offset of the object center relative to the anchor center. 
During the training process, the prediction bounding box whose Intersection over Union (IoU) with the ground truth is greater than 0.7 is taken as the positive sample, 
and the prediction bounding box whose IoU is less than 0.3 is taken as the negative sample. 
During the inference process, the confidence threshold and Non-Maximum Suppression (NMS) \cite{nms} are applied to predict the possible position.

\subsection{Multi-branch orientation estimation network}

\begin{figure}
   % \hspace{0.5cm}
	\centering  %图片全局居中
	% \subfigbottomskip=1pt %两行子图之间的行间距
	% \subfigcapskip=-3pt %设置子图与子标题之间的距离
   \includegraphics[width=0.48\textwidth]{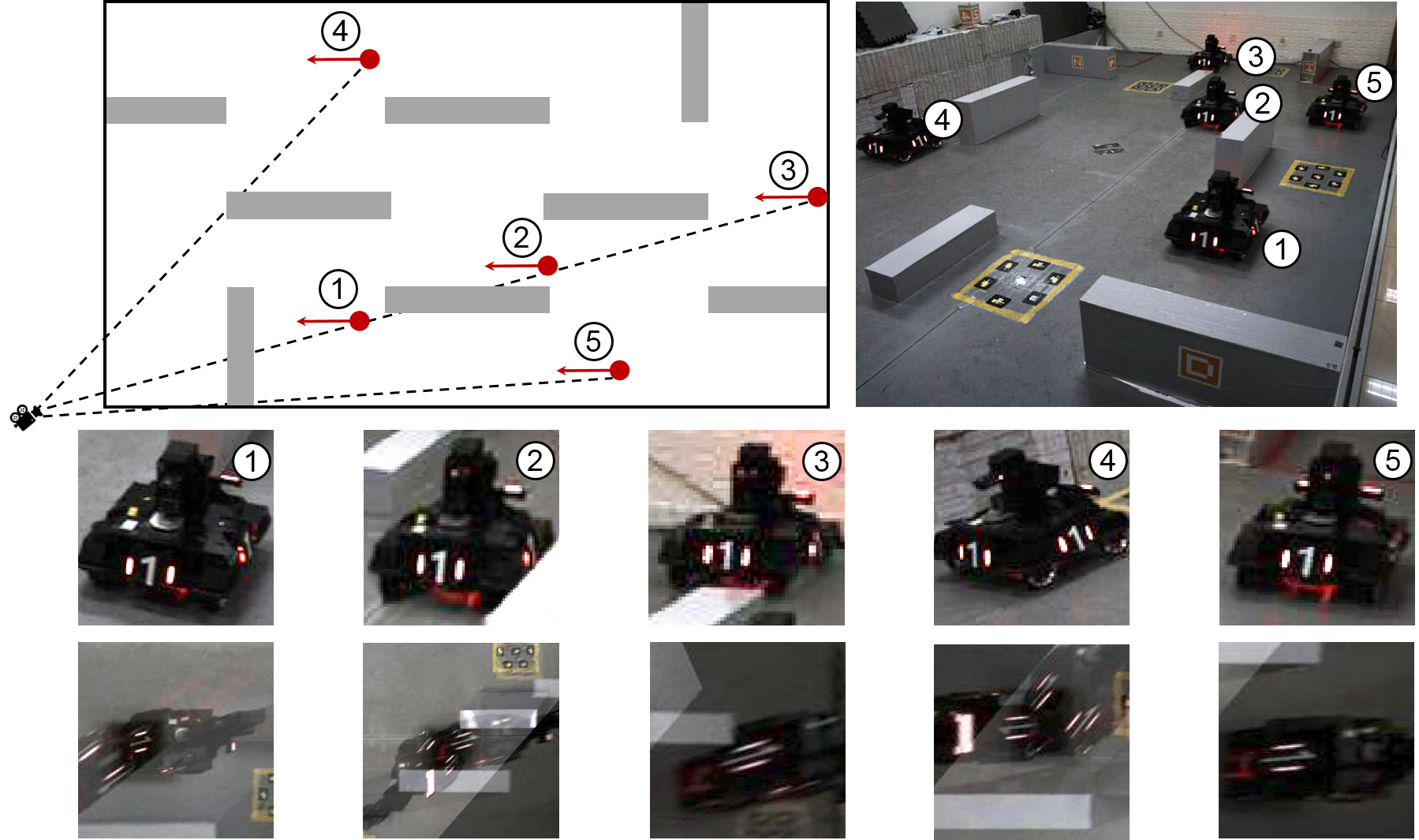} 
   % \begin{minipage}{0.22\textwidth}
	% \includegraphics[width=0.48\textwidth]{imgs/ang5.png} 
	% \includegraphics[width=0.48\textwidth]{imgs/ang6.png} 
   % \center{(a)}
   % \end{minipage}
   % \begin{minipage}{0.22\textwidth}
	% \includegraphics[width=0.48\textwidth]{imgs/ang7.png} 
	% \includegraphics[width=0.48\textwidth]{imgs/ang8.png} 
   % \center{(b)}
   % \end{minipage}
   
	\caption{Confusion problems for orientation estimation. 
   The upper row describes that 5 robots with the same orientation and different positions. 
   The second line is the content of the robots in the image space. The last line is the content of the robots in BEV space.}
   \label{fig_confusion}
   \vspace{-0.5cm}
\end{figure}

Orientation estimation is an essential part of 3D object detection. However, there are several challenges to estimate orientation with multi-view monocular information. 
Similar to the monocular 3D object detection methods, multi-view orientation estimation also faces the same label confusion problem. 
This problem means that the different line of sight relative to the camera makes the observed states various when the object orientation is fixed, 
resulting in the same label corresponding to different states, as shown with \textcircled{2}, \textcircled{4} and \textcircled{5} in Fig. \ref{fig_confusion}. 
The usual solution is to redefine the label according to the line of sight angle to eliminate this ambiguity.
When objects with the same orientation are in the same line of sight, there is no label confusion problem, but due to the influence of the feature orthogonal projection, 
the previous consistent global features face another confusion problem. This is due to the altitude assumption, which causes the streak near the projection center. 
This streak leads to the global features of object at different positions are quite different, even though they have the same orientation on the same line of sight angle. 
This phenomenon is called feature confusion, as shown with \textcircled{1}, \textcircled{2} and \textcircled{3} in Fig. \ref{fig_confusion}.

Since feature confusion is caused by the orthogonal transformation of features, the global features are not suitable for orientation estimation. 
Here we propose {\bf feature perspective pooling} which combines features from each perspective and the position proposals. Firstly, according to the position obtained by PPN, 
the pool region corresponding to the 3D position in the image of each view is calculated through perspective transformation. 
Specifically, 8 vertices of the bounding box are obtained according to the estimated center position of the object and the predefined orientation. 
According to the parameters of the camera, the position of each vertex in the image is calculated by perspective transformation. 
% $$
% U = K [R | T] P
% $$
% where $P$ is the homogeneous coordinates of the vertices. 
We obtain the Region Of Interest (ROI) by calculating the  minimum outer rectangle of the 8 projection vertices.
Then, ROI pooling is used to obtain the features for orientation prediction under each perspective. 
During the perspective pooling process, we assume the altitude and height of the object are known.

Similar to \cite{geometry}, we do not directly regress the orientation, but divide the value into $N$ intervals. 
Each branch network predicts the confidence of the object included in the ROI, the probability that the object angle falls in region $[\frac{2\pi}{N} (i-1), \frac{2 \pi}{N} i], i \in \{1, 2, \cdots, N \}$, 
and the offset of the object angle relative to the center value of the region. During training, the label of offset $o_i$ is calculated as follows

$$
o_i = \beta - \frac{\pi}{N}(2*i -1), \text{where } i \in \{1, 2, \cdots, N \},
$$
where $\beta$ is the ground truth of the orientation. 
In order to eliminate label confusion, the value is the angle relative to the line of sight. 
Since the multi-branch method predicts the orientation of the object in each perspective, each object may have multiple prediction bounding boxes with different orientation. 
During the inference, we first rank the boxes according to the confidence, and then apply NMS to determine the final prediction result.

% The meaning of multi-view pooling based module is to avoid feature confusion comes from feature map projection. When projecting feature maps, even for the same object with the same orientation, objects with different positions in ground plane space will results in different surrounding features, meaning that single label matches multiple feature representation thus causing feature confusion, as illustrated in Figure 3. Therefore, we repreoject 2D birds-eye-view proposals back to origin image feature maps to acquire orientation information. 

% First, we use the pre-defined object demensions $w, l, h$ (respectively are width, length and hight of the robot car in the dataset) to generate 3D boxes. These 3D boxed are represented by 8 points, each point is defined as $[x, y, z]$, where $z = 0$ or $z = h$. Next is to project 3D birds-eye-view points back to origin perspective, for any point $[x, y, z]$ that belongs to 3D proposals, to obtain its corresponding 2D point$[x_{2d}, y_{2d}]$ on origin image, we have
% \begin{equation}
% \begin{bmatrix}
%    x_{2d} \\
%    y_{2d} \\
%    1  \\
% \end{bmatrix}
% = K \cdot [R|t]
% \begin{bmatrix}
%    x\\
%    y\\
%    z\\
%    1
% \end{bmatrix}
% \end{equation}
% where $K$ is a $3 \times 3$ intrinsic matrix, $[R|t]$ is a $3 \times 4$ matrix represents the combination of rotation matrix and translation matrix.

\subsection{Multi-view based 3D object detection}

Based on the PPN and multi-branch orientation estimation network, we propose a multi-view 3D detection method MVM3Det. 
This method takes the monocular image from multiple perspectives as the input and extracts the features through the sharing ResNet-18, 
then obtains the consistent global feature through the feature orthogonal transformation, and applies the anchor based method to estimate the possible spatial position of the object. 
After confidence threshold and NMS, the estimated position is combined with feature perspective pooling to obtain the features of ROI in each perspective and estimates the orientation of the object in each perspective. 
Finally, the bounding boxes are obtained by NMS. In the training process, considering that the orientation estimation network is based on accurate position estimation, 
we adopt a phased training method: first train the PPN, and then train the multi-branch orientation estimation network.

% Based on the proposed modules in previous sections, we build the multi-view based 3D detection network. The key of our algorithm is to produce accurate location information and orientation information. Therefore, our objective is to determine three models: birds-eye-view based region proposal module, multi-view RoI pooling module and multi-view orientation regression module, described as the formula:
% \begin{equation}
%    min {\bm {\mathcal{L}}} (\psi_{r}(S^1,\cdot\cdot\cdot,S^n, \theta)+\psi_{p}(\kappa, \omega) + \psi_{a}(\kappa, \lambda), Gt)   
% \end{equation}
% where $\mathcal{L}$ represents the total loss, $S$ denotes aggregated sensor information, $\psi_{r}$ and $\theta$ represent RPN module and its corresponding weights, $\kappa$ is the proposed regions from RPN module, $\psi_{p}$ and $\omega$ means RoI pooling module and its weights, similarly $\psi_{a}$ and $\lambda$ denotes orientation regression modules and its network parameters.

\subsubsection{Position proposal network loss}
%  To correctly learn the classification ability of recognizing targets and background, given a set of $N$ positive RoI samples $P^{i}$, where $i$ represents the RoI index, we have RPN classification loss as

% PPN predicts the confidence of each anchor containing the object and the offset between the anchor and the ground truth.
% Therefore, the loss function of PPN consists of two parts: confidence loss $\mathcal{L}^{PPN}_{conf}$ and offset loss $\mathcal{L}^{PPN}_{offset}$.

PPN loss $\mathcal{L}^{PPN}$ consists of two parts
 \begin{equation}
   \begin{aligned}
      \mathcal{L}^{PPN} =& \frac{1}{N_{conf}} \sum_i \mathcal{L}_{conf} (\hat{p}_i, p_i) \\
      & + \lambda^{PPN} \frac{1}{N_{val}} \sum_i p_i \mathcal{L}_{offset}(\hat{t}^{BEV}_i, t^{BEV}_i) \\
      & + \lambda^{PPN}_{2D} \frac{1}{N_{val}} \sum_i \sum_v p_i \mathcal{L}_{offset}(\hat{t}^{v}_i, t^{v}_i),
    \end{aligned}
 \end{equation}
where the first part represents the loss of predicted confidence of each anchor.
$\hat{p}_i \in \{0, 1\}$ and $p_i$ are the ground truth label and predicted confidence, respectively. 
$N_{conf}$ represents the total number of anchors.
The second part represents the offset loss of anchor in BEV space, 
$\hat{t}^{BEV}_i$ represents the offset between the true position and anchor, and $t^{BEV}_i$ represents the predicted offset. 
Here, only offset loss of anchor involving objects are considered. $N_{val}$ represents the number of the valide anchors which involve objects.
The last part is multi-brach regression loss which represents the offset loss of 2D bounding box in the image space under the different view $v$.
The box is projected from the BEV position prediction. Experiments show that this loss function improves the accuracy of position estimation.
Softmax loss and smooth $L1$ loss are adopted for $\mathcal{L}_{conf}$ and $\mathcal{L}_{offset}$ respectively.
We follow \cite{fasterrcnn} to encode $t^{BEV}_i$ and $t^{v}_i$. $\lambda^{PPN}$ and $\lambda^{PPN}_{2D}$ are balance parameters.
% where $\hat{p}_i \in \{0, 1\}$ and $p_i$ are the ground truth label and predicted confidence, respectively. 
% $t_i$ includes position offsets and scale offsets between the bounding boxes and anchors. The encoding method is similar to \cite{fasterrcnn}.
% Softmax loss and smooth $L1$ loss are applied to confidence prediction and offset prediction, respectively.

% ---------------------------------------------
\subsubsection{Multi-branch orientation estimation network loss}

For multi-branch orientation estimation network, we predict three parts: 
1)  the confidence of object existence , 2) the orientation interval classification probability  and 3) the orientation offsets.
Therefore, the loss function also consists of three parts.
% alternatively, we combine all the predicted results and calculated only one integrated loss for each task. 
% multi-view classification loss and regression loss are similar to the losses in the position proposal network:
\begin{equation}
   \begin{aligned}
      \mathcal{L}^{MBON} =& \sum_{v} \frac{1}{N^v_{conf}} \sum_i \mathcal{L}_{conf} (\hat{p}^v_i, p^v_i) \\
      & + \frac{1}{N^v_{val}} \sum_i p^v_i \mathcal{L}_{cls} (\hat{b}^v_i, b^v_i) \\
      & + \lambda^{MBON} \frac{1}{N^v_{val}} \sum_i p^v_i \mathcal{L}_{ori}(\hat{o}^v_i, o^v_i)
    \end{aligned}
 \end{equation}
 where $p^v_i$ represents the true label of the $i$-th predicted bounding box in the $v$-th perspective. 
 Second part $\mathcal{L}_{cls}$ is multi-bin classification loss which calculates distribution distance between the one-hot label and predicted probability distribution.
 Here softmax loss is used for this loss.
 $b_i$ represents the probability distribution of the $i$-th box over orientation intervals, 
 and $o_i$ represents the offset between the orientation and the interval center.
 $N^v_{conf}$ is the total number of the predicted bounding boxes in view $v$. $N^v_{val}$ is the number of the validate boxes.
%  Here, softmax loss is used for multibin classification loss $\mathcal{L}_{cls}$. 
 Considering the periodicity of the orientation, we employ the cosine function to encode the offset prediction error as the offset prediction loss $\mathcal{L}_{ori}$, which is same as \cite{geometry}.

\section{EXPERIMENT}
In order to verify the performance of the proposed multi-view 3D detection method, 
we present a multi-view monocular 3D object dataset, and conduct ablation experiments on the dataset. 
In addition, we compare the performance of position detection and orientation estimation with the current SOTA methods on the proposed dataset and public dataset.

% We validate our method on two multi-view detection datasets, the WildTrack pedestrian detection dataset and the RoboMaster dataset. We focus on the localization performance on the WildTrack dataset and both localization and orientation estimation on the RoboMaster dataset.

% \subsection{WildTrack dataset}
% The Wildtrack dataset is a multi-view pedestrian detection dataset proposed by ETH Zurich\cite{wildtrack}. This dataset uses 7 cameras for image capturing. The annotation contains the coordinates of the pedestrians in a fixed plane grid with a size of [480, 1440], the bounding box information in each view, and the calibration parameters for cameras. The average pedestrian density per frame is 20. The dataset contains a total of 400 images and a total of about 56,000 2D bounding boxes. The obstructions are mainly pedestrians.

\subsection{Multi-view monocular 3D object dataset}
Although multi-view 3D object detection is very important for autonomous robot system, there are few relevant algorithms so far, which is largely due to the lack of relevant datasets. 
There are abundant 3D detection datasets for autonomous driving, such as KITTI, Waymo open dataset, etc., 
which contain a large amount of images, point cloud and 3D bounding boxes. However, the data are collected from a single perspective, and there is still a problem of the occlusion. 
The WildTrack dataset is the most similar to our proposed dataset, but it only contains 400 samples and does not provide the orientation label of the object. 
Therefore, it is unable to verify the orientation estimation performance of the methods. 
Our proposed dataset contains 4330 pairs monocular images collected from multiple perspectives and a large number of bounding boxes including position and orientation.
% We create a multi-view 3D detection dataset based on the IEEE ICRA RoboMaster 2021 competition scene to testify our algorithm. 

% The multi-view monocular 3D object dataset is a image dataset for multi-view detection tasks. 
% The data is captured by two cameras and multiple vehicle mounted sensors, and the detection targets are mobile robot cars. 
% In birds-eye-view perspective, we regard the upper left corner as the origin, where right is the positive direction of the x-axis, and bottom is the positive direction of the y-axis. 
% The distances are calculated in centimeters. For birds-eye-view orientation calculation, the right direction is 0 degrees, 
% we define counterclockwise as the positive orientation direction, and the orientation angle is calculated in radians. 
% The annotation contains the coordinates of the robot car in a fixed plane ground with a size of [449, 800], the bounding box information in each perspective, 
% and camera parameters for each camera. The dataset contains a total of 6,900 images and a total of about 46,000 object bounding boxes. 
% The obstructions are mainly walls and bricks in the battleground. We calculate the orientation distribution and occlusion ratio of this dataset, as shown in Table I. 
% We divide $[0,2\pi]$ equally into eight sectors, denoted as Div. 1 to Div. 8, respectively.

The collection of datasets is completed in an 8m$\times$4.5m site. A pair of monocular cameras are erected diagonally on the site to capture images of the site. 
There are some obstacles with different heights and moving objects in the site. In MVM3D dataset, the object to be detected is mobile robot. 
% The main reason for using the robot as the object is that the automatic labeling can be completed with the help of the localization system on the robot, 
% which greatly reduces the manpower required in the manual labeling process. The maximum positioning error of the localization system is less than 10cm. 
% Therefore, the position error of the generated label is not greater than 10cm. 
% In MVM3D, a total of 4330 pairs of images were collected, covering different scenes of one to four robots and various illumination scenes. 
% The right side of Fig. \ref{fig_dataset} shows some samples of the dataset and corresponding labels. Where 3930 pairs are used as the training dataset and 400 pairs are used as the test dataset. 
The orientation distribution of labels in the dataset is very important for orientation estimation. 
% The left part of Fig. \ref{fig_dataset} shows the orientation distribution of the training dataset and the test dataset. 
We try our best to ensure that the number of samples is distributed uniformly in each angle interval.
More information about dataset can be found \url{https://github.com/DRL-CASIA/MVM3D}.

\begin{table}
   \hspace{0cm}
   \center
   \caption{Comparison on WildTrack}
     \begin{tabular}{lcccc}
      \hlinew{1pt}
      \multirow{2}*{Methods} & MODA & MODP & Prec. & Recall\\
     & (\%) & (\%) & (\%) & (\%)\\
     \hline
     RCNN \& clustering \cite{multi-viewrcnn}  & 11.3 & 18.3 & 68 & 43 \\
     POM-CNN \cite{pom2008} & 23.2 & 30.5 & 75 & 55 \\
     DeepMCD \cite{deepmcd} & 67.8  & 64.2 & 85 & 82 \\
     Deep-Occlusion \cite{deepocclusion} & 74.1  & 53.8 & 95 & 80 \\
     MVDet \cite{mvdet} & {\bf 88.2}  & 75.7& {\bf 94.7} & {\bf 93.6} \\
     MVM3Det(Ours) & 84.0  & \bf 75.8 & 93.6 & 90.2 \\
     \hlinew{1pt}
     \end{tabular}%
   
     \label{table_wildtrack}
     \vspace{-0.5cm}
 \end{table}%

% % ---------------------------------
% % ---------------------------------
% ---------------------------------

\subsection{Evaluation metrics}
% Instead of evaluating detected bounding boxes in normal object detection tasks, our method focuses on obtaining global object localization as points. 
% We encode the object localization as $[x_{center}, y_{center}]$, representing the center of the object. 
Following the localization methods such as \cite{mvdet,deepocclusion}, we choose Multi-Object Detection Accuracy (MODA) and Multi-Object Detection Precision (MODP), recall, 
precision as the evaluation metrics for the object localization \cite{moda}. 
% MODA accounts for both false negatives and false positives and is normalized according to the number of frames. 
Average Orientation Similarity (AOS), Average Precision of 3D detection ($AP_{3D}$) \cite{kitti} and Orientation Score (OS) \cite{geometry} 
are selected for the orientation estimation.
% where OS is defined as the ratio of AOS over AP ranging from 0 to 1, describing the similarity only on orientation estimation. 

\subsection{Implementation details}
% During experiments, random brightness, random contrast and random saturation are used for image augmentation. 
% Similar to [10], ResNet-18 is applied as the backbone network to extract deep features from images. 
% The features in BEV space are interpolated into a ﬁxed size [120, 160]. 
% During training PPN, 0.7 and 0.3 are the IoU thresholds and employed to select positive samples and negative samples, respectively. 
% During training multi-branch orientation estimation network, we select 128 samples from PPN with IoU greater than 0.5 to train the network. 
% We use Adam optimizer with learning rate 0.15 × 10 −5 to train the networks for 15 epochs, and batchsize is set to 1.

During experiments, random brightness, random contrast and random saturation are used for image augmentation.
Similar to \cite{mvdet}, ResNet-18 is applied as the backbone network. %  to extract deep features from images
% For experiments on the RoboMaster datasets, we add random image augmentation on input images, the augmentation strategies are $\pm$50\% random brightness, 
% $\pm$50\% random contrast and $\pm$50\% random saturation. 
% We choose to use the convolution layers from Resnet-18 and a self-made fully connected module to replace the origin resnet-18, 
% the fully connected layers are used before RoI Pooling to take the flattened feature maps from Resnet-18 backbone. 
The features in BEV space are interpolated into a fixed size [120, 160]. 
During training PPN, 0.7 and 0.3 are the IoU thresholds and employed to select positive samples and negative sample, respectively.
$\lambda^{PPN}$ is 3, and $\lambda^{PPN}_{2D}$ is 1. 
During training multi-branch orientation estimation network, we select 128 samples from PPN with IoU greater than 0.5 to train the network.
$\lambda^{MBON}$ is 0.4.
% In terms of training RPN network, 
% we set positive sample threshold to 0.8, and set negative threshold to 0.3, maximum number of training samples is 256, and maximum positive ratio is set to 0.8. 
% In terms RoI Pooling, we use RPN to propose at most 2000 candidate anchors and in which we select 128 sample RoIs with IoU greater than 0.5 to train RoI Pooling networks. 
% When projecting the image, we downsample the ground plane by 4$\times$. For joint training, we set $\alpha=3$ to balance loss in formula $(8)$. 
We use Adam optimizer with learning rate $0.15 \times 10 ^ {-5}$ to train the networks for 15 epochs, and batchsize is set to 1. 

 \begin{table}
   \hspace{0cm}
   \center
   \caption{Comparison on MVM3D}
     \begin{tabular}{l c c c c}
      \hlinew{1pt}
      % & \multicolumn{4}{c}{RMMS}\\
   %   \cline{2-5}
   \multirow{2}*{Methods}     & MODA  & MODP  & Prec. & Recall \\
           & (\%)  & (\%)  & (\%) & (\%) \\
     \hline
   %   MVDet(project image) \cite{mvdet}& $68.1^{*}$  & $85.9^{*}$  & $84.0^{*}$  & $84.1^{*}$ \\
   %   MVDet(project results) \cite{mvdet}& $91.9^{*}$  & $88.8^{*}$  & $95.2^{*}$  & $96.8^{*}$\\
   %   MVDet(w/o large kernal) \cite{mvdet}& $87.7^{*}$  & $88.2^{*}$  & $95.1^{*}$  & $92.5^{*}$ \\
     MVDet \cite{mvdet}& \bf $94.8$  & $\bf87.7$  & $97$    & $\bf 98.2$\\
     MVM3Det(Ours) & {\bf 95.9} &  { 83.2}& {\bf 99.2}&  95.8\\
     \hlinew{1pt}
     \end{tabular}%
   \label{table_mvm3d}
   \vspace{-0.5cm}
\end{table}%

\subsection{Localization performance}
% We compare our method with a state-of-the-art method \cite{mvdet} on the WildTrack\cite{wildtrack} dataset and the RoboMaster dataset.
\subsubsection{Results on WildTrack dataset}
On WildTrack dataset, we compare the proposed method with the current SOTA methods, and the results are shown in Table \ref{table_wildtrack}. 
The results show that MVDet is the best positioning method at present. 
Compared with that, the proposed method is slightly better than MVDet in MODP, and the results of other metrics are close. 
Compared with anchor free method MVDet, location estimation of MVM3Det is an anchor based method. 
Due to the limitation of anchor, the recall of this method is lower than that of anchor free method when the objects are dense, so the MODA also is lower.
Compared with the remaining methods, our method has obvious advantages in positioning performance. 
% It is worth mentioning that MVDet does not have the ability of orientation estimation, 
% and our proposed method provides accurate orientation estimation while maintaining the performance of positioning.

\subsubsection{Results on MVM3D dataset}
\begin{figure*}
   % \centering
   % \scalebox{1}{
   % \includegraphics[width=1\textwidth]{imgs/visualization-1.pdf}
   % }
   \begin{minipage}{0.1\textwidth}
      \centering{\bf Ground truth}
   \end{minipage}
   \begin{minipage}{0.9\textwidth}
      \includegraphics[width=\textwidth]{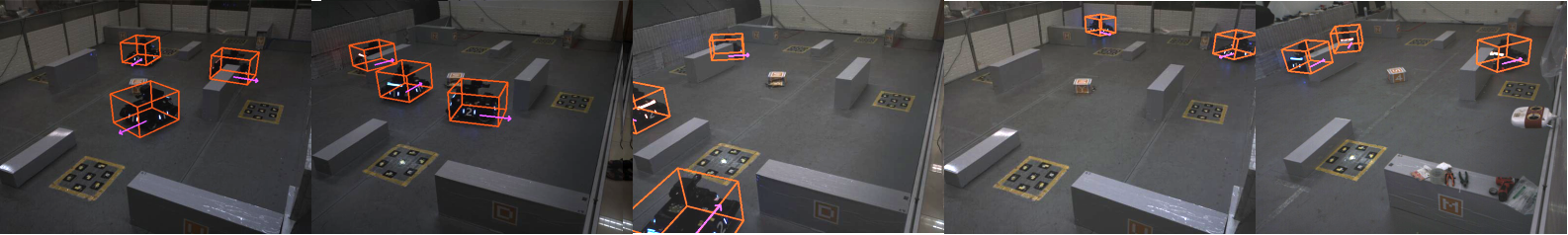}
   \end{minipage}

   \begin{minipage}{0.1\textwidth}
      \centering{\bf MVDet\cite{mvdet}}
   \end{minipage}
   \begin{minipage}{0.9\textwidth}
      \includegraphics[width=\textwidth]{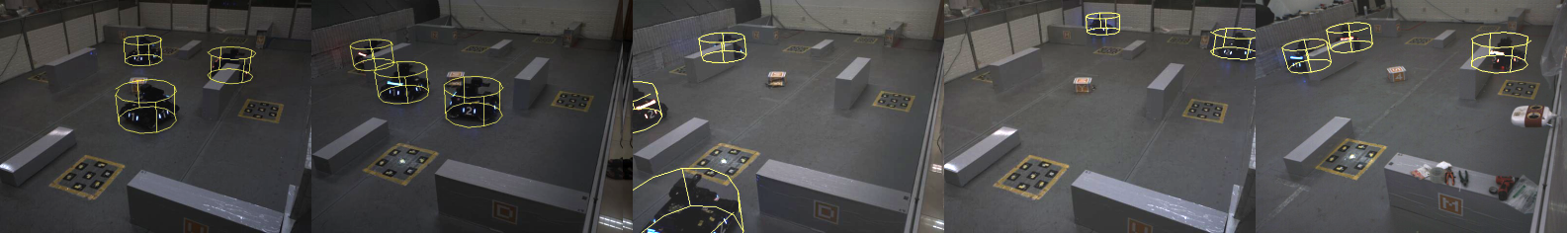}
   \end{minipage}

   \begin{minipage}{0.1\textwidth}
      \centering{\bf MVM3Det}
   \end{minipage}
   \begin{minipage}{0.9\textwidth}
      \includegraphics[width=\textwidth]{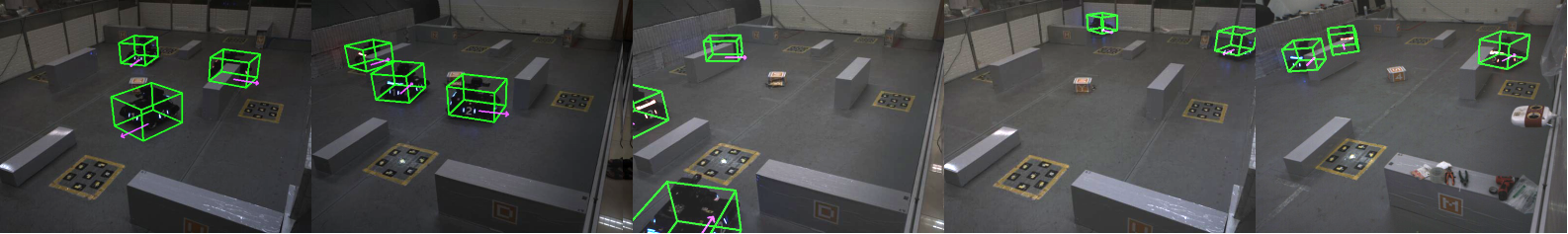}
   \end{minipage}

   \caption{Visualization of the results. Since MVDet does cannot provide orientation information, 
   we generate cylinders according to the predicted location and robots ground truth radius. }
   \label{fig_visualization}
   \vspace{-0.5cm}
\end{figure*}

Compared with WildTrack dataset, MVM3D dataset is larger and the illumination changes are richer. 
Moreover, due to the existence of obstacles, the occlusion situation is more complex. 
% It can be seen from the results in Table \ref{table_mvm3d} that MODA and precision of the proposed method are higher than MVDet. 
It should be noted that although the MVM3D dataset is larger than WildTrack dataset, the density of objects in each frame is less than that of WildTrack. 
This is also the reason why MVM3Det has high recall on this dataset. 
With close recall, anchor based method MVM3Det has better precision, and higher MODA than MVDet, as shown in Table \ref{table_mvm3d}.
The visualization of results on MVM3D dataset are shown in Fig \ref{fig_visualization}. 
It can be seen from the results in the figure that our proposed method can accurately estimate the position and orientation of the object even when it is seriously blocked by obstacles.

% The location estimation module of the proposed method is anchor-based, and its recall depends on the sampling density of anchor. 
% MVDet is an estimation method based on points, which can realize continuous estimation in space. Therefore, when the object distribution density is relatively high, 
% the recall of MVDet method is higher than that of our proposed method.

\begin{table}
   \hspace{0cm}
   \center
   \caption{Ablation studies on MVM3D}
   \scalebox{0.96}{
      \begin{tabular}{ c  c c c  c c c c }
         \hlinew{1pt}
        \multirow{2}*{ } & \multirow{2}*{FOT} & \multirow{2}*{Multi-view}  & \multirow{2}*{MBR}
               &{MODA}   & {MODP}  & {Prec.}  & {Recall}\\
            &  &   &   & (\%) &(\%) & (\%)&  (\%) \\
        \hline
        1 &  &  &  & 12.7 & 43.2 & 50.6 & 22.0\\
        % \h li ne
        2 &\checkmark &  &  & 69.0 & 76.2 & 90.4 & 56.3 \\
        % \hline
        3 &   & \checkmark &    & 27.9 & 69.3 & 91.8 & 30.6 \\
        % \hline
        4  &\checkmark  &\checkmark & & 94.7 & 80.6 & { 96.6} & \bf 99 \\
        5  &\checkmark  &\checkmark  &\checkmark  & {\bf 95.9} &  {\bf 83.2}& \bf  99.2&  { 95.8} \\
        % \hline
        \hlinew{1pt}
        \end{tabular}%
   }
     \label{table_ablation}
     \vspace{-0.5cm}
\end{table}%

\subsection{Ablation studies for localization}

We set up several ablation experiments for the proposed method on MVM3D to analyze the localization performance. The results are shown in Table \ref{table_ablation}. 

\subsubsection{Basic monocular model}
The simplest baseline model is to project the original image into BEV space by feature orthogonal transformation, 
and then uses ResNet-18 as feature extractor to predict the position of objects in BEV space. 
The results are shown in the 1-th row of Table \ref{table_ablation}.

\subsubsection{Feature orthographic transformation (FOT)} 
The baseline model of the 2-th row is modified based on the model of the 1-th row. Specifically, the model applies ResNet-18 to extract features in image space, 
and then uses feature orthogonal transformation to obtain the features in BEV space and predict the spatial position of the object. 
Compared with the results in the 1-th row, it is noticeable that the pre-trained ResNet-18 model has better performance in the original image space.

 \subsubsection{Multi-view image fusion}
 Based on first baseline model, the model in the 3-th row adds the input from multiple perspectives. 
 Images projected from multiple perspectives into the BEV space are stacked as the input to ResNet-18. 
 Compared with the results in the 1-th row, it can be seen that multi-view information increases the location performance to a certain extent.

 \subsubsection{Multi-view feature fusion}
 The model of the 4-th row is based on the 3-th row, which employs ResNet-18 to extract the deep features from the original perspective, 
 and then uses the feature orthogonal transformation to obtain the features in the BEV space. This model is trained with first two part of PPN loss. 
 The results in the 2-th and 4-th rows show that feature fusion under multi-view greatly reduces the false negatives and improves the performance of the method.

 \subsubsection{Multi-branch regression (MBR)}
 The last line is the multi-view 3D object detection method proposed in this paper, which adds a multi-branch 2D bounding box regression loss. 
 Compared with the results in the 4-th and 5-th rows, it can be seen that the joint BEV space loss and multi-branch regress loss increases the position estimation accuracy to a certain extent.

% \subsection{Baslines}
% As our work focuses on addressing occlusion in multi-view detection, we are supposed to compare our methods with other multi-view detection methods. To the best of our knowledge, there are still no multi-view methods similar to ours published. To conduct a thorough comparison of our multi-view method.
% We evaluate localization performance on the RoboMaster and Wildtrack datasets and evaluate multi-view orientation estimation on the RoboMaster datasets. For localization performance, we choose to compare with the state-of-the-art RGB image-based multi-view detection method MVDet\cite{mvdet}. For orientation estimation, we compare our method with our own baseline method.
% ---------------------------------
 % ---------------------------------

 \subsection{Orientation performance}

 \begin{table}
   \hspace{0cm}
   \center
   \caption{Orientation performance on MVM3D}
     \begin{tabular}{cccc|ccc}
      \hlinew{1pt}
      \multirow{2}*{Methods} & \multicolumn{3}{c|}{IoU = 0.25} & \multicolumn{3}{c}{IoU = 0.5} \\
     & $AP_{3D}$ & AOS& OS & $AP_{3D}$ & AOS & OS \\
     \hline
      Monocular& 77.2\% & 67.1\% & 0.87 & 30.3\% & 26.6\% & 0.88 \\
      MVM3Det & {\bf 90.2\%}  & {\bf 82.6\%}& {\bf 0.91} & \bf 49.0\% &  \bf 45.5\% & \bf 0.92 \\
     \hlinew{1pt}
     \end{tabular}%
   \label{table_orientation}
   \vspace{-0.5cm}
 \end{table}%

 As far as we know, there is no method to realize orientation estimation with multi-view monocular information. 
 However, orientation is necessary to monocular 3D detection. 
%  We introduce AOS and OS as the evaluation metrics of orientation estimation. 
%  AP is a comprehensive metric for monocular 3D detection which evaluates position and orientation estimation simultaneously. 
 The baseline model compared here is a degraded model based on our proposed method, which only uses single view information. 
 It can be seen from the results in Table \ref{table_orientation} that multi-view fusion greatly improves the accuracy of orientation estimation.

\section{CONCLUSIONS}
In this paper, a novel multi-view monocular 3D object detection method is proposed, 
which overcomes false negatives caused by the occlusion and the confusion problem of orientation estimation under multiple perspectives. 
This method is mainly composed of PPN and multi-branch orientation estimation network. 
PPN fuses the data from different perspectives through the feature orthogonal transformation to estimate the spatial position of the objects. 
Multi-branch orientation estimation introduces feature perspective pooling to realize orientation estimation from various perspectives, 
so as to alleviate the problems of label confusion and feature confusion in orientation estimation. In addition, in order to promote the development of multi-view 3D object detection, 
we present the first multi-view monocular 3D dataset MVM3D, covering different illumination and complex occlusion scenes. Through the experiments on our dataset and public dataset, 
the proposed method achieves competitive results to the current SOTA in position estimation, and realizes orientation estimation for the first time.
% \section{ACKNOWLEDGEMENTS}
% test test test

{
\bibliographystyle{IEEEtran}
\bibliography{IEEEroot}
}

\end{document}